\documentclass[]{fairmeta}
\usepackage{mathpazo}
\usepackage{tgpagella}

\usepackage{xcolor}
\usepackage{amsmath}
\usepackage{amssymb}
\usepackage{graphicx}
\usepackage{arydshln}
\usepackage{multirow}
\usepackage{booktabs}
\usepackage{marvosym}  
\usepackage{colortbl}
\usepackage{float}
\usepackage{cuted}  
\usepackage{pifont}
\usepackage{pgfplots}
\usepackage{wrapfig}

\newcommand{\cmark}{\ding{51}}%
\newcommand{\xmark}{\ding{55}}%

\def \ie {\emph{i.e.}}
\def \eg {\emph{e.g.}}
\def \etc {\emph{etc.}}

\definecolor{darkgreen}{rgb}{0.04,0.63,0.07}
\definecolor{skyblue}{rgb}{0.04,0.40,0.80}
\definecolor{tablegray}{gray}{.9}

\newcommand{\ours}{\textit{Leffa}}

\title{Learning Flow Fields in Attention for Controllable Person Image Generation}

\author[1,2,*]{Zijian Zhou}
\author[1]{Shikun Liu}
\author[1]{Xiao Han}
\author[1]{Haozhe Liu}
\author[1]{Kam Woh Ng}
\author[1]{Tian Xie}
\author[1]{Yuren Cong}
\author[1]{Hang Li}
\author[1]{Mengmeng Xu}
\author[1]{Juan-Manuel Pérez-Rúa}
\author[1]{Aditya Patel}
\author[1]{Tao Xiang}
\author[3]{Miaojing Shi}
\author[1]{Sen He}

\affiliation[1]{Meta AI}
\affiliation[2]{King's College London}
\affiliation[3]{Tongji University}

\contribution[*]{Work done at Meta}

\date{\today}

\abstract{
Controllable person image generation aims to generate a person image conditioned on reference images, allowing precise control over the person’s appearance or pose.
However, prior methods often distort fine-grained textural details from the reference image, despite achieving high overall image quality.
We attribute these distortions to inadequate attention to corresponding regions in the reference image.
To address this, we thereby propose \textbf{learning flow fields in attention} (\textbf{\ours{}}), which explicitly guides the target query to attend to the correct reference key in the attention layer during training.
Specifically, it is realized via a regularization loss on top of the attention map within a diffusion-based baseline.
Our extensive experiments show that \ours{} achieves state-of-the-art performance in controlling appearance (virtual try-on) and pose (pose transfer), significantly reducing fine-grained detail distortion while maintaining high image quality.
Additionally, we show that our loss is model-agnostic and can be used to improve the performance of other diffusion models.

}

\begin{document}

\maketitle

\begin{figure}[h]
    \centering
    \vspace{-2mm}
    \includegraphics[width=\textwidth]{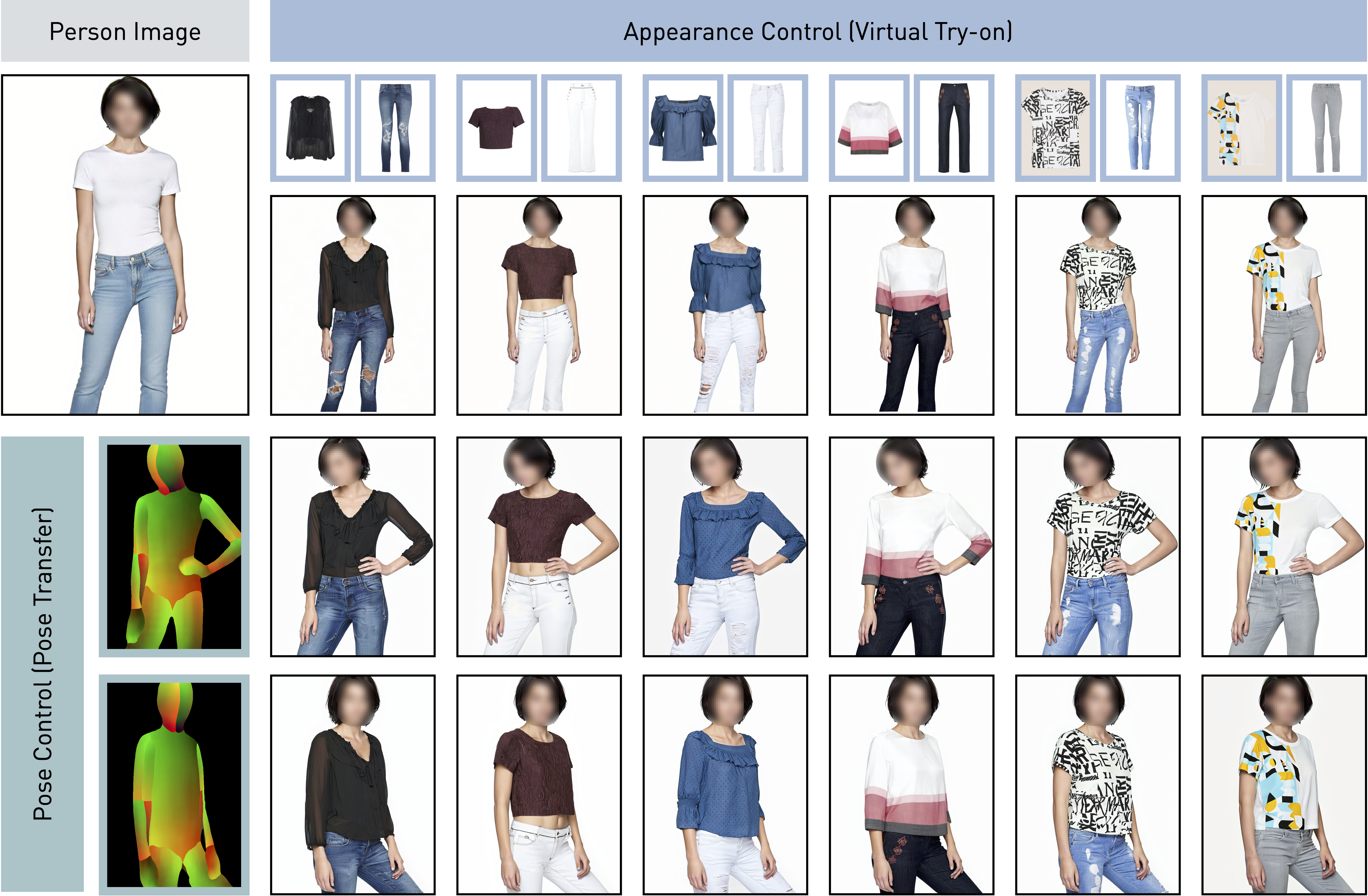}
    \vspace{-6mm}
    \caption{
    We present \ours{}, a unified framework for controllable person image generation that enables precise manipulation of both appearance (\ie, virtual try-on) and pose (\ie, pose transfer). \ours{}'s generated images demonstrate high quality, with fine details preserved and minimal texture distortion.
    Please zoom in for better viewing.
    }
    \vspace{-6mm}
\end{figure}

\section{Introduction}
\label{sec:introduction}

Controllable person image generation aims to generate an image of a person guided by reference images, allowing for control over attributes such as appearance and pose.
Underpinning a wide variety of applications in virtual and augmented reality, gaming, and e-commerce, this task has made significant progress~\citep{kim2024stableviton, choi2024improving, chong2024catvton, wan2024improving, bhunia2023person, han2023controllable, lu2024coarse, pham2024cross} based on the advancements in diffusion-based image generation~\citep{ho2020denoising, rombach2022high}.

\begin{wrapfigure}{r}{0.6\textwidth}
    \vspace{-4mm}
    \centering
    \begin{subfigure}[t]{0.145\textwidth}
        \centering
        \includegraphics[width=\textwidth]{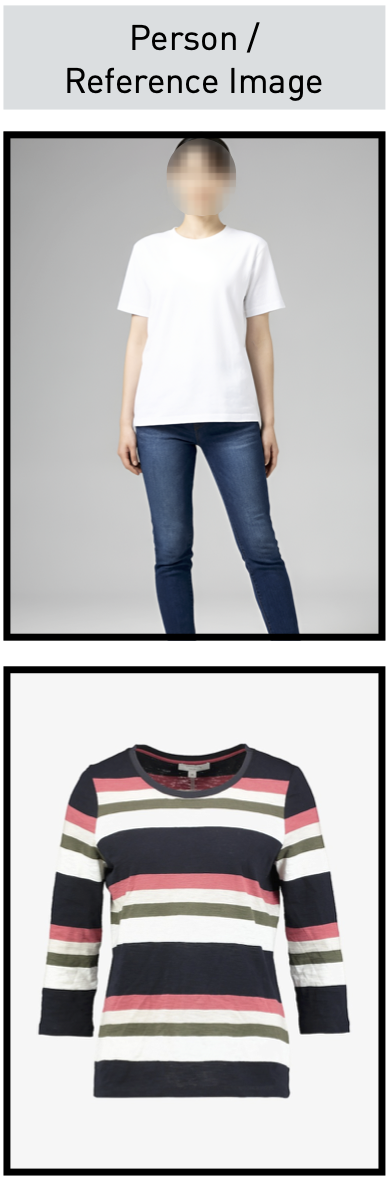}
        \vspace{-5mm}
        \caption{}
        \label{subfig:intro_1}
    \end{subfigure}\hspace{0mm}
    \begin{subfigure}[t]{0.145\textwidth}
        \centering
        \includegraphics[width=\textwidth]{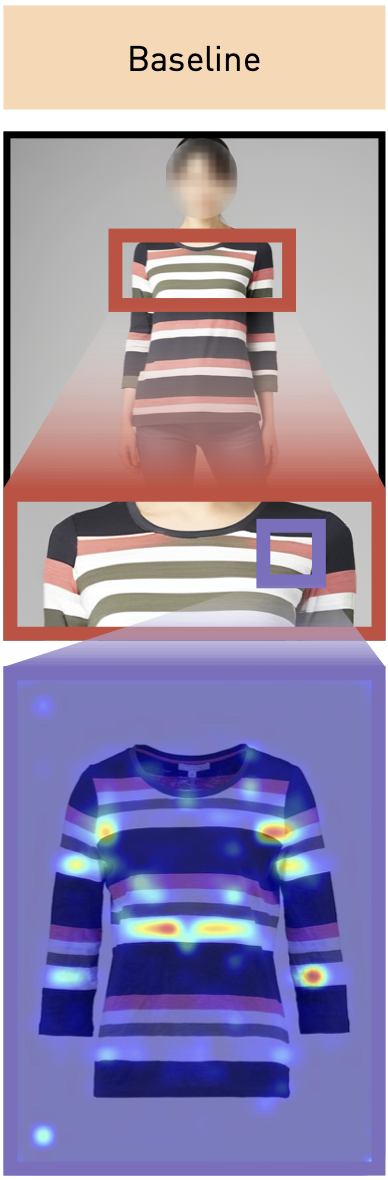}
        \vspace{-5mm}
        \caption{}
        \label{subfig:intro_2}
    \end{subfigure}\hspace{0mm}
    \begin{subfigure}[t]{0.145\textwidth}
        \centering
        \includegraphics[width=\textwidth]{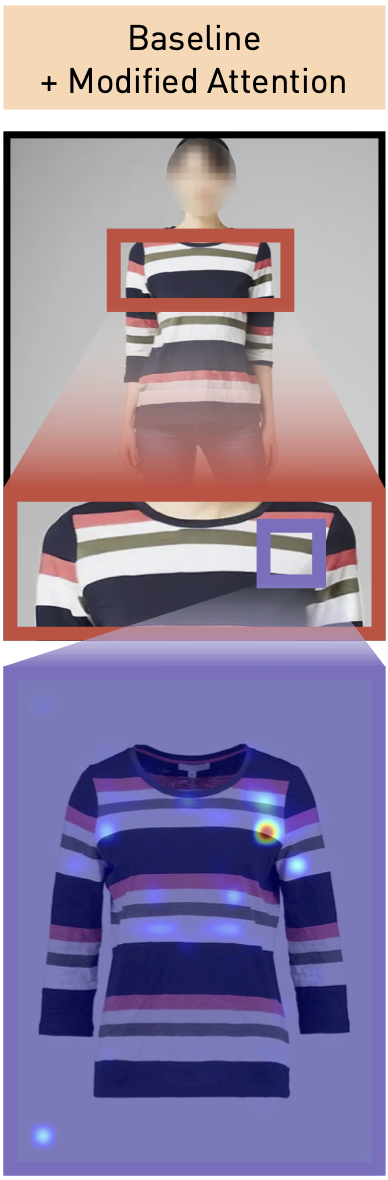}
        \vspace{-5mm}
        \caption{}
        \label{subfig:intro_3}
    \end{subfigure}\hspace{0mm}
    \begin{subfigure}[t]{0.145\textwidth}
        \centering
        \includegraphics[width=\textwidth]{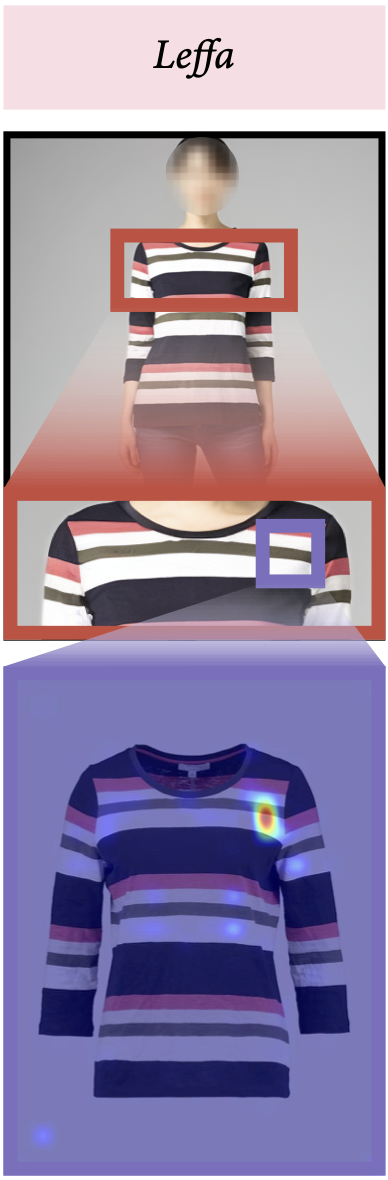}
        \vspace{-5mm}
        \caption{}
        \label{subfig:intro_4}
    \end{subfigure}
    \vspace{-3mm}
    \caption{
    Taking appearance control of a person image (virtual try-on) as an example:
    (a) input person and reference (garment) images;
    (b) generated image and attention map from a diffusion-based method (\eg, IDM-VTON~\citep{choi2024improving});
    (c) generated image after \textit{manually modifying} the attention map in the diffusion-based method to focus on the correct regions;
    (d) generated image and attention map from \ours{}.
    Our method generates high-quality images without detail distortion (see the colored striped texture).
    }
    \vspace{-4mm}
    \label{fig:introduction}
\end{wrapfigure}

The primary challenge in this task is to preserve the \textit{fine-grained details} (\eg, textures, texts, logos, \etc) from the reference image while maintaining \textit{overall quality}.
Although current methods can generate high-quality images at first glance, they still suffer from the distortion of fine-grained textural details upon closer inspection (see Fig.~\ref{subfig:intro_2} top, where the striped texture is wrong).
Researchers have attempted to improve the preservation of fine-grained details by integrating inference strategies (\eg, disentangled guidance~\citep{bhunia2023person}, multi-stage inference~\citep{yang2024texture}, incorporating information from other modalities~\citep{ning2024picture, xu2024ootdiffusion, choi2024improving} (\eg, textual descriptions) and introducing auxiliary models for feature improvement~\citep{kim2024stableviton, xu2024ootdiffusion, choi2024improving, wan2024improving, sun2024outfitanyone, lu2024coarse, pham2024cross} (\eg, using CLIP~\citep{radford2021learning}, DINOv2~\citep{oquab2023dinov2} features, and warping model, \etc).
However, these methods typically increase model complexity and lack \textit{explicit} supervision to establish visual consistencies between the target and reference images.

By visualizing the attention layer of various diffusion-based methods~\citep{choi2024improving, xu2024ootdiffusion}, we observe that in regions where details are distorted, the target query exhibits \textit{widely-distributed} attention rather than accurately focusing on the corresponding reference key regions.
This observation also aligns with findings from previous studies~\citep{kim2024stableviton, ren2022neural}.
For instance, in Fig.~\ref{subfig:intro_2}, the incorrect generation of striped texture results from the target query failing to attend to the corresponding reference key regions.
In contrast, when the attention map is \textit{manually corrected} by swapping the highest-response values with those lying in the correct region, the model is able to generate significantly improved striped textures without any additional training (see Fig.~\ref{subfig:intro_3}).

Capitalizing on this observation, we propose a regularization loss by \textbf{le}arning \textbf{f}low \textbf{f}ields in \textbf{a}ttention (\textbf{\ours{}}) to alleviate the distortion of fine-grained details.
Specifically, we transform the attention map between the target query and reference key into a flow field during training, which warps the reference image to more closely align with the target image. This process encourages the target query to accurately attend to the correct reference key.

To validate our method, we incorporate our \ours{} loss into a diffusion-based baseline, observing a significant reduction in fine-grained detail distortion while maintaining high image quality (see Fig.~\ref{subfig:intro_4} top), without additional inference cost.
Additionally, visualization of our method's attention map (see Fig.~\ref{subfig:intro_4} bottom) confirms the model accurately focuses on the desired regions.
We quantitatively and qualitatively evaluate our method on virtual try-on (appearance control) with VITON-HD~\citep{choi2021viton} and DressCode~\citep{morelli2022dress}; and pose transfer (pose control) with DeepFashion~\citep{liu2016deepfashion}, both achieving state-of-the-art performance and mitigating the distortion of fine-grained details across all datasets.
Finally, we show that our \ours{} loss is model-agnostic and generalizes effectively to other diffusion-based methods~\citep{choi2024improving, chong2024catvton}, demonstrating \ours{} as a versatile and general solution for controllable person image generation.

The contributions of this work are as follows:
\begin{itemize}
    \item We propose \ours{}, a regularization loss that explicitly guides the target queries in attention layers to focus on correct reference keys, mitigating detail distortion without additional parameters and inference cost. 
    \item We evaluate \ours{} on both virtual try-on and pose transfer across various diffusion-based methods, achieving state-of-the-art quantitative performance and significantly reducing fine-grained detail distortion qualitatively.
\end{itemize}

\section{Preliminary}
\label{sec:preliminary}

\subsection{Diffusion Model}

Diffusion models~\citep{ho2020denoising, rombach2022high} have gained popularity in image generation models due to their stable training and exceptional generative capabilities.
Among them, the Stable Diffusion (SD) series (\eg, SD1.5~\citep{rombach2022high}, SDXL~\citep{podell2023sdxl}) is widely used.
SD is a latent diffusion model (LDM), consisting of a variational autoencoder (VAE)~\citep{kingma2013auto}, a UNet~\citep{ronneberger2015u} $\epsilon_{\theta}(\cdot)$, and text encoders~\citep{raffel2020exploring, radford2021learning}.
The VAE includes an encoder $\mathcal{E}(\cdot)$ and a decoder $\mathcal{D}(\cdot)$, which encodes input images from a pixel space into a latent space to reduce computational cost.
During training, we first use the encoder $\mathcal{E}(\cdot)$ of VAE to encode a given image into a latent feature $\mathbf{z}_\text{0}$.
At a sampled diffusion time step $t \in \{1, \cdots, T\}$, we apply a Gaussian noise $\epsilon \sim \mathcal{N}(0, 1)$ to the latent feature $\mathbf{z}_\text{0}$ based on a noise scheduler, obtaining a noised latent feature $\textbf{z}_t$.
The UNet $\epsilon_{\theta}(\cdot)$ then denoises the noised latent feature $\textbf{z}_t$ under the control of the conditional feature $\textbf{c}$.
Formally, the training loss can be formulated as:
\begin{equation}
    \mathcal{L}_{\text{diffusion}} = \mathbb{E}_{\epsilon \sim \mathcal{N}(0,1),t} [ \| \epsilon - \epsilon_{\theta}(\textbf{z}_t, \textbf{c}, t) \|_{2}^{2} ],
\label{eq:loss_diffusion}
\end{equation}
where $\textbf{c}$ can be either the language feature predicted by the text encoders or the visual feature derived from the reference image, which is the primary focus of our work.

\subsection{Task Definition and Notations}

\noindent Given a source person image $I_{\text{src}} \in \mathbb{R}^{H \times W \times 3}$ and a reference image $I_{\text{ref}} \in \mathbb{R}^{H \times W \times 3}$, controllable person image generation aims to generate a target person image $I_{\text{tgt}} \in \mathbb{R}^{H \times W \times 3}$, where $H$ and $W$ denote the height and width of the image.
Specifically, in virtual try-on, $I_{\text{tgt}}$ represents the garment image $I_{\text{ref}}$ worn on the person in $I_{\text{src}}$; while in pose transfer, $I_{\text{tgt}}$ represents the person in $I_{\text{ref}}$ adopting the pose from $I_{\text{src}}$.
During training, $I_{\text{src}}$ and $I_{\text{tgt}}$ are the same image.
The data preprocessing for each task is defined as follows:

\begin{itemize}
    \item For virtual try-on, following~\cite{choi2024improving, chong2024catvton}, we first extract a garment mask $I_\text{m} \in \{0, 1\}^{H \times W \times 1}$ from $I_\text{src}$ and construct a garment-agnostic image $I_\text{ga} \in \mathbb{R}^{H \times W \times 3} = I_\text{src} * (1 - I_\text{m})$.
    During training, we encode $I_\text{src}$ and $I_\text{ga}$ via the VAE encoder $\mathcal{E}(\cdot)$ yields $z_\text{0}$ and $z_\text{ga}$, and resize $I_\text{m}$ to latent space as $z_\text{m}$.
    We add noise to $z_\text{0}$ based on the sampled time step $t$ which produces $z_t$, and we channel concatenate it with $z_\text{ga}$ and $z_\text{m}$ to construct $\hat{z_t}$.
    \item For pose transfer, following~\cite{han2023controllable}, we first extract a DensePose image~\citep{guler2018densepose} $I_\text{dp} \in \mathbb{R}^{H \times W \times 3}$ from $I_\text{src}$.
    During training, we encode $I_\text{src}$ with the VAE encoder $\mathcal{E}(\cdot)$, obtaining $z_\text{0}$, and resize the $I_\text{dp}$ to the latent resolution of $z_\text{dp}$.
    We add noise to $z_\text{0}$ at sampled time step $t$ to produce $z_t$, and we channel concatenate it with $z_\text{dp}$ to construct $\hat{z_t}$.
\end{itemize}
For the reference image $I_{\text{ref}}$ in both tasks, we encode it into a latent feature $z_{\text{ref}}$, and use that as a conditional feature within a diffusion loss defined in Eq.~\ref{eq:loss_diffusion}.

\begin{figure*}[t]
    \centering
    \includegraphics[width=1.0\linewidth]{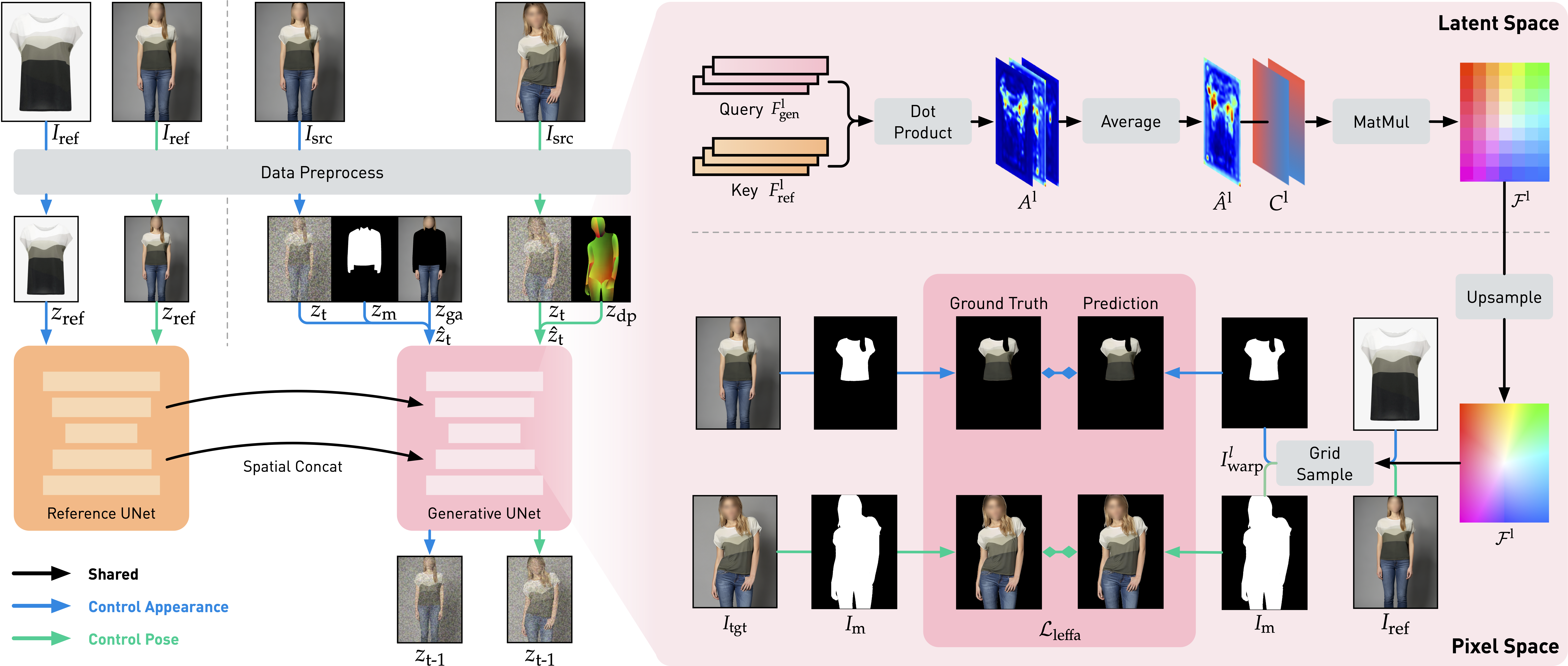}
    \vspace{-6mm}
    \caption{An overview of our \ours{} training pipeline for controllable person image generation.
    The left is our diffusion-based baseline; the right is our \ours{} loss.
    Note that $I_\text{src}$ and $I_\text{tgt}$ are the same image during training.
    }
    \label{fig:overview}
\end{figure*}

\section{Method}
\label{sec:method}

Our goal is to leverage diffusion-based methods to preserve fine-grained details while maintaining high overall image quality.
To achieve this, we propose the \ours{} loss that alleviates detail distortion through learning flow fields in attention.
To validate the effectiveness of \ours{}, we first construct a unified diffusion-based baseline suitable for both virtual try-on and pose transfer tasks in Sec.~\ref{sec:baseline}.
We then introduce the formulation of \ours{} loss in Sec.~\ref{sec:leffa}.
Lastly, we explain its integration into the model training in Sec.~\ref{sec:training}.

\subsection{Diffusion-based Baseline}
\label{sec:baseline}

We first design our diffusion-based baseline upon the SD~\citep{rombach2022high, podell2023sdxl} to assess the effectiveness of our \ours{} loss.
Specifically, we modify SD in two steps as follows.

First, we duplicate the pre-trained SD UNet $\epsilon_{\theta}(\cdot)$ to create two separate UNets: i) a Generative UNet $\epsilon_{\theta_{\text{gen}}}(\cdot)$ for processing $\hat{z_t}$ from the source image $I_\text{src}$; and ii) a Reference UNet $\epsilon_{\theta_{\text{ref}}}(\cdot)$ for processing $z_\text{ref}$ from the reference image $I_\text{ref}$.
Since we only condition on images, we remove both the text encoders and the cross-attention layers used for text interaction in the Generative UNet $\epsilon_{\theta_{\text{gen}}}(\cdot)$ and Reference UNet $\epsilon_{\theta_{\text{ref}}}(\cdot)$.
Notably, both UNets are \textit{fully trainable}, which we found to be effective in our early exploration.

Second, to condition the generation of the target person image $I_{\text{tgt}}$ on $I_{\text{src}}$ and $I_{\text{ref}}$, we employ \textit{spatial concatenated} self-attention.
For $l$-th attention layer, we define the generative features fed into the Generative UNet $\epsilon_{\theta_{\text{gen}}}(\cdot)$ as $F_{\text{gen}}^{l} \in \mathbb{R}^{n^{l} \times d^{l}}$, and the reference features in the Reference UNet $\epsilon_{\theta_{\text{ref}}}(\cdot)$ as $F_{\text{ref}}^{l} \in \mathbb{R}^{n^{l} \times d^{l}}$, where $n^{l} = h^{l} \times w^{l}$ with $h^{l}$ and $w^{l}$ as spatial dimensions, and $d^{l}$ as the feature dimension.
We concatenate the $F_{\text{ref}}^{l}$ with $F_{\text{gen}}^{l}$ along the spatial dimension, yielding concatenated feature $F_{\text{cat}}^{l} \in \mathbb{R}^{2n^{l} \times d^{l}}$, and jointly feed it into the Generative UNet's self-attention layer.
We retain only the first half of the output corresponding to $F_{\text{gen}}^{l}$ for further processing.
This process is repeated across all self-attention layers.

Using SD's training method and the loss function in Eq.~\ref{eq:loss_diffusion}, our clean and simple diffusion-based baseline already achieves performance comparable to current state-of-the-arts~\citep{kim2024stableviton, choi2024improving, chong2024catvton, wan2024improving, bhunia2023person, han2023controllable, lu2024coarse, pham2024cross}, while relying on no additional auxliary models (\eg, CLIP~\citep{radford2021learning}, DINOv2~\citep{oquab2023dinov2}) and/or complicated training techniques (\eg, DREAM~\citep{zhou2024dream}).
However, this diffusion-based baseline still suffers from detail distortion, which will be addressed by the \ours{} loss, introduced in the next section.

\subsection{\textbf{\ours{}}: Learning Flow Fields in Attention}
\label{sec:leffa}

Our goal is to preserve fine-grained details without textural distortion.
Detail distortion arises when the target query in the attention layer fails to attend to the corresponding region of the reference key correctly.
To address this, we propose the \ours{} loss, which explicitly guides the target query to \textit{spatially focus} on the correct region of the reference key through supervision by learning flow fields in attention.

Specifically, in the $l^{th}$ attention layer of the Generative UNet $\epsilon_{\theta_{\text{gen}}}(\cdot)$, we compute the attention map $A^{l}$ by using $F_{\text{gen}}^{l}$ as the query $Q$ and $F_{\text{ref}}^{l}$ as the key $K$, formulated as the dot-product attention defined in \cite{vaswani2017attention},
\begin{equation}
    A^{l} = \texttt{softmax} \bigg( \frac{Q K^{\top}}{\sqrt{d}} / \tau \bigg),
\end{equation}
where $\tau$ is the temperature coefficient, and $d$ is the token size of the query $Q$.
We then construct the \ours{} loss based on the attention map $A^{l}$.

For each attention map $A^{l}$, we aim to encourage all query tokens in $F_{\text{gen}}^{l}$ to attend to the correct regions on the reference key $F_{\text{ref}}^{l}$.
Following \citet{darcet2023vision}, we also include additional learnable tokens to allow flexible attention on other regions.
This design, rather than enforcing strict spatial attention alignment in all attention heads, we ensure that, \textit{on average}, attention maps across different heads focus on the correct region.
To achieve this, we average $A^{l}$ across the head dimension, resulting in $\hat{A^{l}} \in \mathbb{R}^{n^{l} \times n^{l}}$.

Next, we convert $\hat{A^{l}}$ to a flow field representing the spatial correspondence between $I_\text{src}$ and $I_\text{ref}$.
To achieve this, we construct a normalized coordinate map $C^{l} \in \mathbb{R}^{n^{l} \times 2}$, where, reshaped as $\mathbb{R}^{h^{l} \times w^{l} \times 2}$, within the top-left coordinate $(0, 0)$ has the value of $[-1, -1]$ and the bottom-right coordinate $(h-1, w-1)$ has the value of $[1, 1]$.
We then multiply the attention map $\hat{A^{l}}$ with the coordinate map $C^{l}$ to obtain the flow field $\mathcal{F}^{l} \in \mathbb{R}^{n^{l} \times 2}$, formulated as
\begin{equation}
    \mathcal{F}^{l} = \hat{A^{l}} \cdot C^{l},
\end{equation}
which indicates the coordinates of the region on the reference feature $F_{\text{ref}}^{l}$ that each target query token attends to.

We then use this flow field $\mathcal{F}^{l}$ as a coordinate mapping and apply the grid sampling operation to warp the reference image $I_{\text{ref}}$ into the target image $I_{\text{tgt}}$.
Considering our method is based on the latent diffusion model, the resolution of the attention map is significantly lower than the original input image.
To provide precise pixel-level training supervision, we upsample the flow field in $l^{th}$ attention layer $\mathcal{F}^{l}\in\mathbb{R}^{h^{l} \times w^{l} \times 2}$ to the original image resolution using bilinear interpolation, resulting in the upsampled flow field $\mathcal{F}_{up}^{l} \in \mathbb{R}^{H \times W \times 2}$.
We then use this upsampled flow field $\mathcal{F}_{up}^{l}$ to warp the reference image $I_{\text{ref}}$, producing the warped image ${I_{\text{warp}}^{l}} \in \mathbb{R}^{H \times W \times 3}$.

Finally, we compute the L2 loss between ${I_{\text{warp}}^{l}}$ and the corresponding region of the target image $I_{\text{tgt}}$, which defines our \ours{} loss formulation:
\begin{equation}
     \mathcal{L}_{\text{leffa}} = \sum\nolimits_{l=1}^{L}  \Vert I_{\text{tgt}} * I_m - {I_{\text{warp}}^{l}} * I_m \Vert_{2}^{2}.
\end{equation}
This loss supervises the attention map without additional inputs or parameters, ensuring each target query token attends to the correct reference region and preserving fine-grained detail consistency in the generated images.

Note that, we apply \ours{} loss only on selected $L$ attention layers, explained next.

\subsection{Model Training with \textbf{\ours{}} Loss}
\label{sec:training}
In this section, we detail the application of \ours{} loss to the diffusion-based baseline, guiding the target query to attend to the correct reference key and thus alleviating fine-grained detail distortion.

We follow \citet{zhu2024m} to adopt progressive training with \ours{} loss, applied in the final finetuning phase to avoid early-stage performance degradation.
The model is initially trained at a low resolution (\eg, $256$, $512$), then at a higher resolution (\eg, $1024$).
In the final phase, we fine-tune at $1024$ resolution using a combined loss:
\begin{equation}
    \mathcal{L}_{\text{finetune}} = \mathcal{L}_{\text{diffusion}} + \lambda_\text{leffa} \mathcal{L}_{\text{leffa}},
\end{equation}
where $\lambda_\text{leffa}$ is the loss weight of $\mathcal{L}_{\text{leffa}}$.
For optimal model performance with \ours{} loss, we also address the following considerations.

\noindent \textbf{Attention Layer Selection.}
Given that current LDM-based methods perform feature interaction between the target and reference images in the latent space, the resulting attention maps tend to have relatively low resolution.
For such low-resolution attention maps, accurately warping the reference image $I_{\text{ref}}$ is not feasible.
Therefore, we set a resolution threshold $\theta_{\text{resolution}} = h/H$ (\eg, $1/32$), representing the ratio of the attention map size to the original image size (see Fig.~\ref{fig:ablation}).
Only attention layers with a resolution greater than $\theta_{\text{resolution}}$ participate in the computation of $\mathcal{L}_{\text{leffa}}$.

\noindent \textbf{Timestep Selection.}
When the time step $t$ is large, substantial noise is added to the image, making accurate feature interactions between the target and reference difficult due to the excessive noise, which hinders attending to the right semantics of the corresponding regions.
Therefore, \ours{} loss is not suitable for large time steps (see Fig.~\ref{fig:ablation}).
We thereby set a timestep threshold $\theta_{\text{timestep}}$ (\eg, $500$ when $T=1000$), and during training, only time steps smaller than $\theta_{\text{timestep}}$ are included in the calculation of $\mathcal{L}_{\text{leffa}}$.

\noindent \textbf{Temperature Selection.}
Different temperatures adjust the smoothness of the attention map produced by softmax.
A larger temperature results in a smoother attention map, allowing the target query token to attend to a broader range of reference features $F_{\text{ref}}^{l}$, which helps the model more easily learn the correct reference regions and increases tolerance to errors (see Fig.~\ref{fig:ablation}).
Therefore, when using \ours{} loss, we apply a relatively large temperature $\tau$ (\ie, 2.0).

\begin{figure*}[t]
    \centering
    \includegraphics[width=1.0\linewidth]{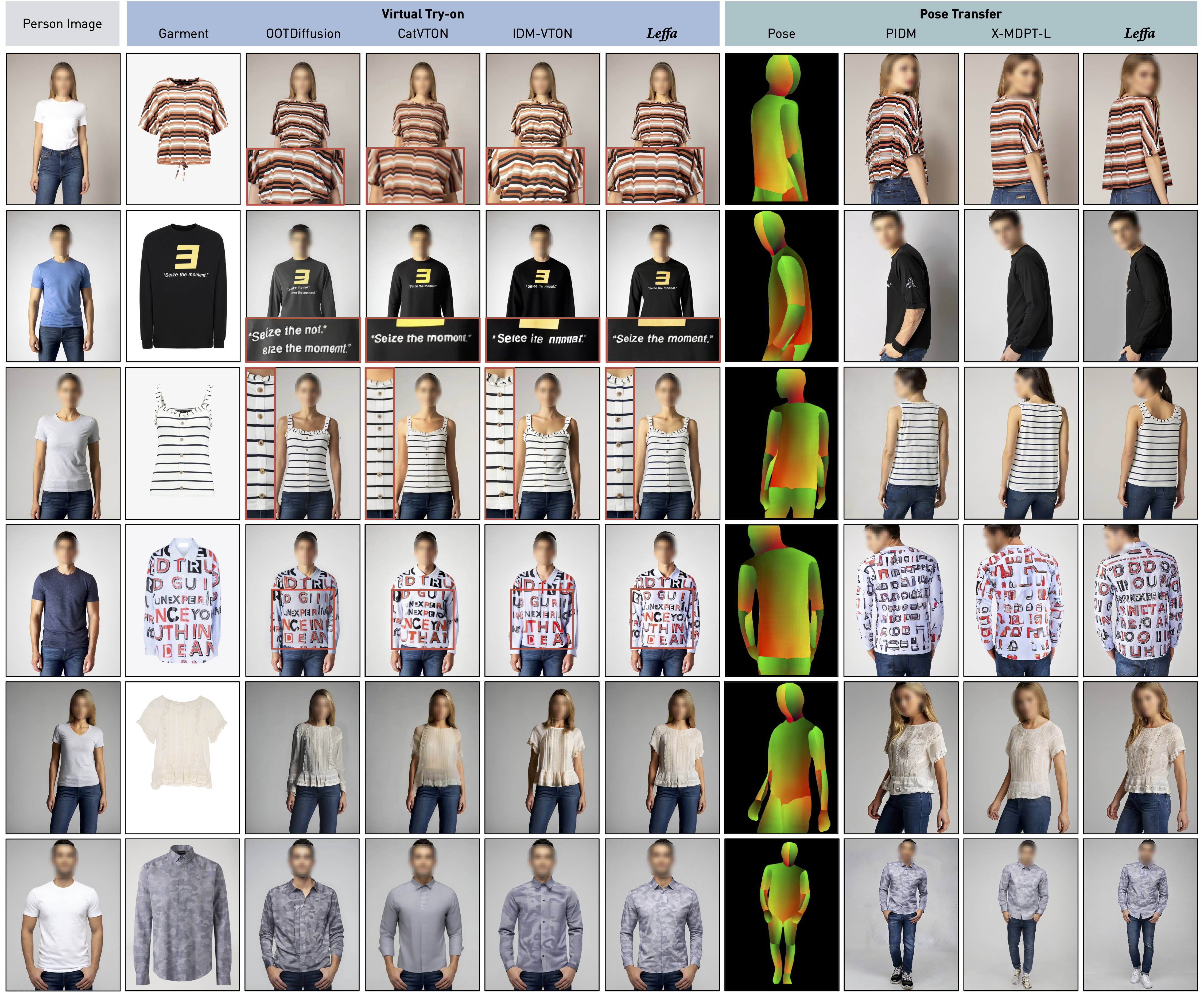}
    \vspace{-8mm}
    \caption{
    Qualitative visual results comparison with other methods.
    The input person image for the pose transfer is generated using our method in the virtual try-on.
    The visualization results demonstrate that our method not only generates high-quality images but also greatly reduces the distortion of fine-grained details.
    Please zoom in for better viewing.
    }
    \label{fig:vis}
    \vspace{-4mm}
\end{figure*}

\section{Experiments}
\label{sec:exps}

\subsection{Datasets, Metrics and Implementation Details}
\label{sec:datasets}

\noindent \textbf{Datasets.}
To evaluate our method's capability in appearance and pose control, we conduct experiments on the virtual try-on and pose transfer tasks, respectively.
For virtual try-on, we use the VITON-HD~\citep{choi2021viton} and DressCode~\citep{morelli2022dress} datasets; for pose transfer, we use DeepFashion~\citep{liu2016deepfashion}. Please refer to the supplementary material for more details.

\noindent \textbf{Metrics.}
In virtual try-on, following previous methods~\citep{zhu2023tryondiffusion, gou2023taming, cui2023learning, cui2023street, zeng2024cat, ning2024picture, kim2024stableviton, xu2024ootdiffusion, choi2024improving, yang2024texture, chong2024catvton, yang2024d}, we evaluate the model under two settings: \textit{paired}, where the garment in the person image same as the garment image; and \textit{unpaired}, where the garments differ between the two images.
For both settings, we use FID~\citep{soloveitchik2021conditional} and KID~\citep{binkowski2018demystifying}
as evaluation metrics.
Additionally in the paired setting, we include SSIM~\citep{wang2004image} and LPIPS~\citep{zhang2018unreasonable} to evaluate image-level metrics with ground-truth data.
In pose transfer, we follow the previous methods~\citep{bhunia2023person, han2023controllable, lu2024coarse, pham2024cross} to use FID~\citep{soloveitchik2021conditional}, calculated between the generated test images and all training images; SSIM~\citep{wang2004image} and LPIPS~\citep{zhang2018unreasonable}, computed between the generated images and all images in the test set, as metrics.

\noindent \textbf{Implementation Details.}
In this paper, we use SD1.5~\citep{rombach2022high} to build our diffusion-based baseline.
We adopt the progressive training strategy to train the model.
For virtual try-on, the model is first trained at $512 \times 384$ resolution for $10\text{k}$ steps with a batch size of 256, then at $1024 \times 768$ for $36\text{k}$ steps with a batch size of 64.
Finally, we finetune the model for an additional $10k$ steps, incorporating $\mathcal{L}_{\text{leffa}}$ to enhance its capabilities.
For pose transfer, we start at a resolution of $256 \times 176$ with a batch size of 256 for $80\text{k}$ steps, then increase to $512 \times 352$ 
for another $80\text{k}$ steps.
Finally, we finetune for $24\text{k}$ steps with $\mathcal{L}_{\text{leffa}}$.
AdamW optimizer~\citep{loshchilov2017decoupled} with a learning rate of $10^{-5}$ is applied to all training stages.
The hyperparameters for $\mathcal{L}_{\text{leffa}}$ are set as follows: $\lambda_{\text{leffa}} = 10^{-3}$, $\theta_{\text{resolution}} = 1/32$, $\theta_{\text{timestep}} = 500$ and $\tau = 2.0$.

\subsection{Quantitative and Qualitative Results}
\label{sec:main_results}

\noindent \textbf{Quantitative Comparison.}
We conduct experiments on both virtual try-on and pose transfer tasks.
For \textit{virtual try-on}, we conduct experiments on VITON-HD~\citep{choi2021viton} and DressCode~\citep{morelli2022dress}, with results shown in Tabs.~\ref{tab:viton_hd} and \ref{tab:dresscode}, respectively.
In VITON-HD, our method significantly outperforms previous methods, achieving an FID reductions of -0.88/-0.5 in the paired/unpaired setting.
In DressCode, our method also surpasses existing methods across all garment categories (upper body, lower body, and dresses), with an FID reductions of -1.93/-1.66 in the paired/unpaired setting.
We also conduct evaluations for each garment category, please refer to the supplementary material.
For \textit{pose transfer}, we conduct experiments on DeepFashion~\citep{liu2016deepfashion} at different resolutions, significantly outperforming previous methods.
As shown in Tab.~\ref{tab:deepfashion}, our method achieves -0.54 and -1.61 reductions in FID at $256 \times 176$ and $512 \times 352$, respectively.
While our model shows a minor SSIM gap over CFLD~\citep{lu2024coarse}, visual results and human study indicate substantial improvement over prior methods (see Fig.~\ref{fig:human_study} and supplementary material).

\begin{table}[!t]
\centering
\resizebox{0.75\linewidth}{!}{
\begin{tabular}{lcccccc}
\toprule[1.5pt]
\multirow{2}{*}{\textbf{Method}}        & \multicolumn{4}{c}{\textbf{paired}} & \multicolumn{2}{c}{\textbf{unpaired}} \\ \cmidrule{2-7} 
                                        & \textbf{FID} $\downarrow$ & \textbf{KID} $\downarrow$ & \textbf{SSIM} $\uparrow$ & \textbf{LPIPS} $\downarrow$ & \textbf{FID} $\downarrow$ & \textbf{KID} $\downarrow$ \\ \midrule
FS-VTON~\citep{he2022style}             & 6.17  & 0.69 & 0.886 & 0.074 & 9.91  & 1.10 \\
HR-VITON~\citep{lee2022high}            & 11.38 & 3.52 & 0.865 & 0.122 & 13.27 & 4.38 \\
GP-VTON~\citep{xie2023gp}               & 6.03  & 0.60 & 0.885 & 0.080 & 9.37  & 0.79 \\
LADI-VTON~\citep{morelli2023ladi}       & 6.60  & 1.09 & 0.866 & 0.094 & 9.35  & 1.66 \\
DCI-VTON~\citep{gou2023taming}          & 5.52  & \underline{0.41} & 0.882 & 0.080 & 8.75  & 0.68 \\
SD-VTON~\citep{shim2024towards}         & 6.98  & 1.00 & 0.874 & 0.101 & 9.85  & 1.39 \\
StableVITON~\citep{kim2024stableviton}  & 8.23  & 0.49 & \underline{0.888} & 0.073 & -     & -    \\
IDM-VTON~\citep{choi2024improving}      & 5.76  & 0.73 & 0.850 & 0.063 & 9.84  & 1.12 \\
OOTDiffusion~\citep{xu2024ootdiffusion} & 9.30  & 4.09 & 0.819 & 0.088 & 12.41 & 4.68 \\
CatVTON~\citep{chong2024catvton}        & \underline{5.42}  & \underline{0.41} & 0.870 & \underline{0.057} & \underline{9.02}  & \underline{1.09} \\
\textbf{\ours{} (Ours)}                 & \textbf{4.54}  & \textbf{0.05} & \textbf{0.899} & \textbf{0.048} & \textbf{8.52}  & \textbf{0.32} \\
\bottomrule[1.5pt]
\end{tabular}
}
\vspace{-2mm}
\caption{Quantitative results comparison with other methods on the VITON-HD dataset for virtual try-on. Note \textbf{bold} indicates the best result, and \underline{underline} indicates the second-best result.}
\label{tab:viton_hd}
\end{table}

\begin{table}[!t]
\centering
\resizebox{0.75\linewidth}{!}{
\begin{tabular}{lcccccc}
\toprule[1.5pt]
\multirow{2}{*}{\textbf{Method}}        & \multicolumn{4}{c}{\textbf{paired}} & \multicolumn{2}{c}{\textbf{unpaired}} \\ \cmidrule{2-7} 
                                        & \textbf{FID} $\downarrow$ & \textbf{KID} $\downarrow$ & \textbf{SSIM} $\uparrow$ & \textbf{LPIPS} $\downarrow$ & \textbf{FID} $\downarrow$ & \textbf{KID} $\downarrow$ \\ \midrule
GP-VTON~\citep{xie2023gp}               & 9.93  & 4.61 & 0.771 & 0.180 & 12.79 & 6.63 \\
LADI-VTON~\citep{morelli2023ladi}       & 9.56  & 4.68 & 0.766 & 0.237 & 10.68 & 5.79 \\
IDM-VTON~\citep{choi2024improving}      & 6.82  & 2.92 & 0.879 & 0.056 & 9.55  & 4.32 \\
OOTDiffusion~\citep{xu2024ootdiffusion} & 4.61  & 0.96 & 0.885 & 0.053 & 12.57 & 6.63 \\
CatVTON~\citep{chong2024catvton}        & \underline{3.99}  & \underline{0.82} & \underline{0.892} & \underline{0.045} & \underline{6.14}  & \underline{1.40} \\
\textbf{\ours{} (Ours)}                 & \textbf{2.06}  & \textbf{0.07} & \textbf{0.924} & \textbf{0.031} & \textbf{4.48} & \textbf{0.62} \\
\bottomrule[1.5pt]
\end{tabular}
}
\vspace{-2mm}
\caption{Quantitative results comparison with other methods on the DressCode dataset (all garment categories) for virtual try-on. 
}
\vspace{-4mm}
\label{tab:dresscode}
\end{table}

\noindent \textbf{Qualitative Comparison.}
Fig.~\ref{fig:vis} shows that our method significantly improves detail preservation compared to previous methods.
For \textit{virtual try-on}, in the first row, our method accurately preserves the texture details of the colored stripes and maintains the correct color order.
In the second row, it generates very small text accurately, whereas other methods, despite achieving reasonable overall quality, exhibit varying levels of text distortion, failing to convey the correct meaning.
In the third row, our method generates evenly spaced buttons in the correct quantity, and in the fifth row, it produces a more realistic, gauzy fabric effect.
For \textit{pose transfer}, in the first to fourth rows, our method better predicts a plausible side/back-view image based on frontal information.
Other rows also show that our method maintains fine details more effectively when altering the pose.
The last row demonstrates that our method can generate vivid full-body images from half-body images.

\begin{table}[!t]
\setlength{\tabcolsep}{0.4cm}
\centering
\resizebox{0.75\linewidth}{!}{
\begin{tabular}{clccc}
\toprule[1.5pt]
\textbf{Resolution} & \textbf{Method} & \textbf{FID} $\downarrow$ & \textbf{SSIM} $\uparrow$ & \textbf{LPIPS} $\downarrow$ \\
\midrule
\multirow{13}{*}{$256 \times 176$}  & Def-GAN \citep{siarohin2018deformable} & 18.46 & 0.679 & 0.233 \\
                                    & PATN \citep{zhu2019progressive}        & 20.75 & 0.671 & 0.256 \\
                                    & ADGAN \citep{men2020controllable}      & 14.46 & 0.672 & 0.228 \\
                                    & GFLA \citep{ren2020deep}               & 10.57 & 0.707 & 0.223 \\
                                    & PISE \citep{zhang2021pise}             & 13.61 & 0.663 & 0.206 \\
                                    & DPTN \citep{zhang2022exploring}        & 11.39 & 0.711 & 0.193 \\
                                    & CASD \citep{zhou2022cross}             & 11.37 & 0.725 & 0.194 \\
                                    & NTED \citep{ren2022neural}             & 8.68  & 0.718 & 0.175 \\
                                    & PIDM \citep{bhunia2023person}          & 6.37  & 0.731 & 0.168 \\
                                    & PoCoLD \citep{han2023controllable}     & 8.07  & 0.731 & \underline{0.164} \\
                                    & CFLD \citep{lu2024coarse}              & 6.80  & \textbf{0.737} & 0.152 \\
                                    & X-MDPT-L \citep{pham2024cross}         & \underline{6.25}  & 0.729 & 0.167 \\
                                    & \textbf{\ours{} (Ours)}               & \textbf{5.71}  &
                                    \underline{0.732} & \textbf{0.114} \\
\midrule
\multirow{8}{*}{$512 \times 352$}   & CocosNet-v2 \citep{zhou2021cocosnet}   & 13.33 & 0.724 & 0.226 \\
                                    & NTED \citep{ren2022neural}             & 7.78  & 0.738 & 0.198 \\
                                    & PIDM \citep{bhunia2023person}          & \underline{5.84}  & 0.742 & 0.177 \\
                                    & PoCoLD \citep{han2023controllable}     & 8.42  & 0.743 & 0.192 \\
                                    & CFLD \citep{lu2024coarse}              & 7.14  & \underline{0.748} & \underline{0.152} \\
                                    & X-MDPT-L \citep{pham2024cross}         & 5.93  & 0.742 & 0.179 \\
                                    & \textbf{\ours{} (Ours)} & \textbf{4.23}  & \textbf{0.755} & \textbf{0.119} \\
\bottomrule[1.5pt]
\end{tabular}
}
\vspace{-2mm}
\caption{Quantitative results comparison with other methods on the DeepFashion dataset for pose transfer. 
}
\vspace{-2mm}
\label{tab:deepfashion}
\end{table}

\subsection{Generalization to Other Diffusion Models}

The proposed \ours{} loss is model-agnostic and can be applied to different attention-based methods.
We evaluate this by adding \ours{} loss to two prior art diffusion-based methods (\ie, IDM-VTON~\citep{choi2024improving} and CatVTON~\citep{chong2024catvton}) on the VITON-HD~\citep{choi2021viton} dataset.
Since both methods also use attention mechanism to interact with the reference image, our \ours{} loss can be seamlessly integrated without introducing extra trainable parameters.
As shown in Tab.~\ref{tab:general}, our method achieves FID reductions of -0.64/-0.6 for IDM-VTON and -0.56/-0.25 for CatVTON in paired/unpaired settings, confirming its generalization ability.

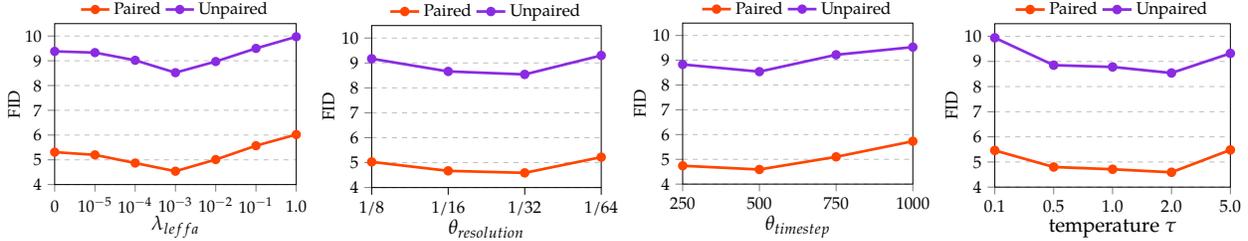
\begin{figure*}[!t]
    \centering
    \begin{subfigure}{0.251\textwidth}
        \centering
        \resizebox{\linewidth}{!}{{
\begin{tikzpicture}

\definecolor{optiona}{RGB}{255,69,0}
\definecolor{optionb}{RGB}{138,43,226}

\begin{axis}[
width=1.61\textwidth,
height=1.2\textwidth,
ymajorgrids,
grid style=dashed,
legend columns=-1,
legend cell align={left},
legend style={
    fill=none,
    draw opacity=1,
    text opacity=1,
    at={(0.5,1.18)},
    anchor=north,
    draw=none,
    very thick,
}, 
tick align=outside,
tick pos=left,
x grid style={darkgray176},
xlabel={\large $\lambda_{leffa}$},
xmin=0, xmax=6,
xtick={0,1,2,3,4,5,6},
xticklabels={
  $0\vphantom{^{0}}$, $10^{-5}$,$10^{-4}$, $10^{-3}$, $10^{-2}$, $10^{-1}$,  $1.0\vphantom{^{0}}$
},
ylabel={FID},
ymin=4, ymax=10.5,
ytick={4,5,6,7,8,9,10},
y label style={at={(axis description cs:0.12,.5)}, anchor=south},
]
\addplot [ultra thick, optiona, mark=*, mark size=2, mark options={solid,rotate=180}]
table {%
0 5.31
1 5.2
2 4.87
3 4.54
4 5.01
5 5.57
6 6.02
};
\addlegendentry{Paired}
\addplot [ultra thick, optionb, mark=*, mark size=2, mark options={solid,rotate=180}]
table {%
0 9.38
1 9.33
2 9.02
3 8.52
4 8.97
5 9.5
6 9.97
};
\addlegendentry{Unpaired}
\end{axis}

\end{tikzpicture}}}
        \label{subfig:ablation_1}
    \end{subfigure}\hfill
    \begin{subfigure}{0.249\textwidth}
        \centering
        \resizebox{\linewidth}{!}{{
\begin{tikzpicture}

\definecolor{optiona}{RGB}{255,69,0}
\definecolor{optionb}{RGB}{138,43,226}

\begin{axis}[
width=1.54\textwidth,
height=1.2\textwidth,
ymajorgrids,
grid style=dashed,
legend columns=-1,
legend cell align={left},
legend style={
    fill=none,
    draw opacity=1,
    text opacity=1,
    at={(0.5,1.18)},
    anchor=north,
    draw=none,
    very thick,
}, 
tick align=outside,
tick pos=left,
x grid style={darkgray176},
xlabel={\large $\theta_{resolution}$},
xmin=0, xmax=3,
xtick style={color=black},
xtick={0,1,2,3},
xticklabels={1/8,1/16,1/32,1/64},
ylabel={FID},
ymin=4, ymax=10.5,
ytick={4,5,6,7,8,9,10},
y label style={at={(axis description cs:0.12,.5)}, anchor=south},
]
\addplot [ultra thick, optiona, mark=*, mark size=2, mark options={solid,rotate=180}]
table {%
0 5.03
1 4.67
2 4.59
3 5.22
};
\addlegendentry{Paired}
\addplot [ultra thick, optionb, mark=*, mark size=2, mark options={solid,rotate=180}]
table {%
0 9.17
1 8.66
2 8.54
3 9.3
};
\addlegendentry{Unpaired}
\end{axis}

\end{tikzpicture}}}
        \label{subfig:ablation_2}
    \end{subfigure}\hfill
    \begin{subfigure}{0.249\textwidth}
        \centering
        \resizebox{\linewidth}{!}{{
\begin{tikzpicture}

\definecolor{optiona}{RGB}{255,69,0}
\definecolor{optionb}{RGB}{138,43,226}

\begin{axis}[
width=1.55\textwidth,
height=1.2\textwidth,
ymajorgrids,
grid style=dashed,
legend columns=-1,
legend cell align={left},
legend style={
    fill=none,
    draw opacity=1,
    text opacity=1,
    at={(0.5,1.18)},
    anchor=north,
    draw=none,
    very thick,
}, 
tick align=outside,
tick pos=left,
xlabel={\large $\theta_{timestep}$},
xmin=0, xmax=3,
xtick={0,1,2,3},
xticklabels={250,500,750,1000},
ylabel={FID},
ymin=4, ymax=10.5,
ytick={4,5,6,7,8,9,10},
y label style={at={(axis description cs:0.12,.5)}, anchor=south},
]
\addplot [ultra thick, optiona, mark=*, mark size=2, mark options={solid,rotate=180}]
table {%
0 4.74
1 4.59
2 5.1
3 5.73
};
\addlegendentry{Paired}
\addplot [ultra thick, optionb, mark=*, mark size=2, mark options={solid,rotate=180}]
table {%
0 8.83
1 8.54
2 9.22
3 9.53
};
\addlegendentry{Unpaired}
\end{axis}

\end{tikzpicture}}}
        \label{subfig:ablation_3}
    \end{subfigure}\hfill
    \begin{subfigure}{0.249\textwidth}
        \centering
        \resizebox{\linewidth}{!}{{
\begin{tikzpicture}

\definecolor{optiona}{RGB}{255,69,0}
\definecolor{optionb}{RGB}{138,43,226}

\begin{axis}[
width=1.56\textwidth,
height=1.2\textwidth,
ymajorgrids,
grid style=dashed,
legend columns=-1,
legend cell align={left},
legend style={
    fill=none,
    draw opacity=1,
    text opacity=1,
    at={(0.5,1.18)},
    anchor=north,
    draw=none,
    very thick,
}, 
tick align=outside,
tick pos=left,
x grid style={darkgray176},
xlabel={\large temperature $\tau$},
xmin=0, xmax=4,
xtick={0,1,2,3,4},
xticklabels={0.1,0.5,1.0,2.0,5.0},
ylabel={FID},
ymin=4, ymax=10.5,
ytick={4,5,6,7,8,9,10},
y label style={at={(axis description cs:0.12,.5)}, anchor=south},
]
\addplot [ultra thick, optiona, mark=*, mark size=2, mark options={solid,rotate=180}]
table {%
0 5.46
1 4.8
2 4.71
3 4.59
4 5.48
};
\addlegendentry{Paired}
\addplot [ultra thick, optionb, mark=*, mark size=2, mark options={solid,rotate=180}]
table {%
0 9.94
1 8.85
2 8.78
3 8.54
4 9.32
};
\addlegendentry{Unpaired}
\end{axis}

\end{tikzpicture}}}
        \label{subfig:ablation_4}
    \end{subfigure}
    \vspace{-12mm}
    \caption{Ablation study for (a) \ours{} loss weight $\lambda_{\text{leffa}}$, (b) resolution threshold $\theta_{\text{resolution}}$, (c) timestep threshold $\theta_{\text{timestep}}$, (d) temperature $\tau$ on VITON-HD dataset of virtual try-on.}
    \vspace{-2mm}
    \label{fig:ablation}
\end{figure*}

\begin{table}[!t]
\setlength{\tabcolsep}{0.4cm}
\centering
\resizebox{0.75\linewidth}{!}{
\begin{tabular}{lccccccc}
\toprule[1.5pt]
\multirow{2}{*}{\textbf{Method}}  & \multirow{2}{*}{\textbf{$\mathcal{L}_{\text{leffa}}$}}        & \multicolumn{4}{c}{\textbf{paired}} & \multicolumn{2}{c}{\textbf{unpaired}} \\ \cmidrule{3-8} 
                                        & & \textbf{FID} $\downarrow$ & \textbf{KID} $\downarrow$ & \textbf{SSIM} $\uparrow$ & \textbf{LPIPS} $\downarrow$ & \textbf{FID} $\downarrow$ & \textbf{KID} $\downarrow$ \\ 
\midrule
\multirow{2}{*}{IDM-VTON}   & \xmark          & 5.76  & 0.73 & 0.850 & 0.063 & 9.84  & 1.12 \\ 
  & \cmark & 5.12 & 0.35 & 0.876 & 0.054 & 9.24 & 0.87 \\ \midrule[0.5pt]
\multirow{2}{*}{CatVTON}   & \xmark          & 5.42  & 0.41 & 0.870 & 0.057 & 9.02  & 1.09 \\
   & \cmark & 4.86 & 0.28 & 0.886 & 0.056 & 8.77 & 0.58 \\ \midrule[0.5pt]
\multirow{2}{*}{Ours}     & \xmark          & 5.31  & 0.30 & 0.885 & 0.058 & 9.38  & 0.91 \\
      & \cmark & \textbf{4.54}  & \textbf{0.05} & \textbf{0.899} & \textbf{0.048} & \textbf{8.52}  & \textbf{0.32} \\
\bottomrule[1.5pt]
\end{tabular}
}
\vspace{-2mm}
\caption{Applying our \ours{} loss $\mathcal{L}_\text{leffa}$ consistently enhances performance across different diffusion-based methods.}
\vspace{-2mm}
\label{tab:general}
\end{table}

\subsection{Ablation Study}
\label{sec:ablation}

\begin{figure}[t]
    \centering
    \includegraphics[width=1.0\linewidth]{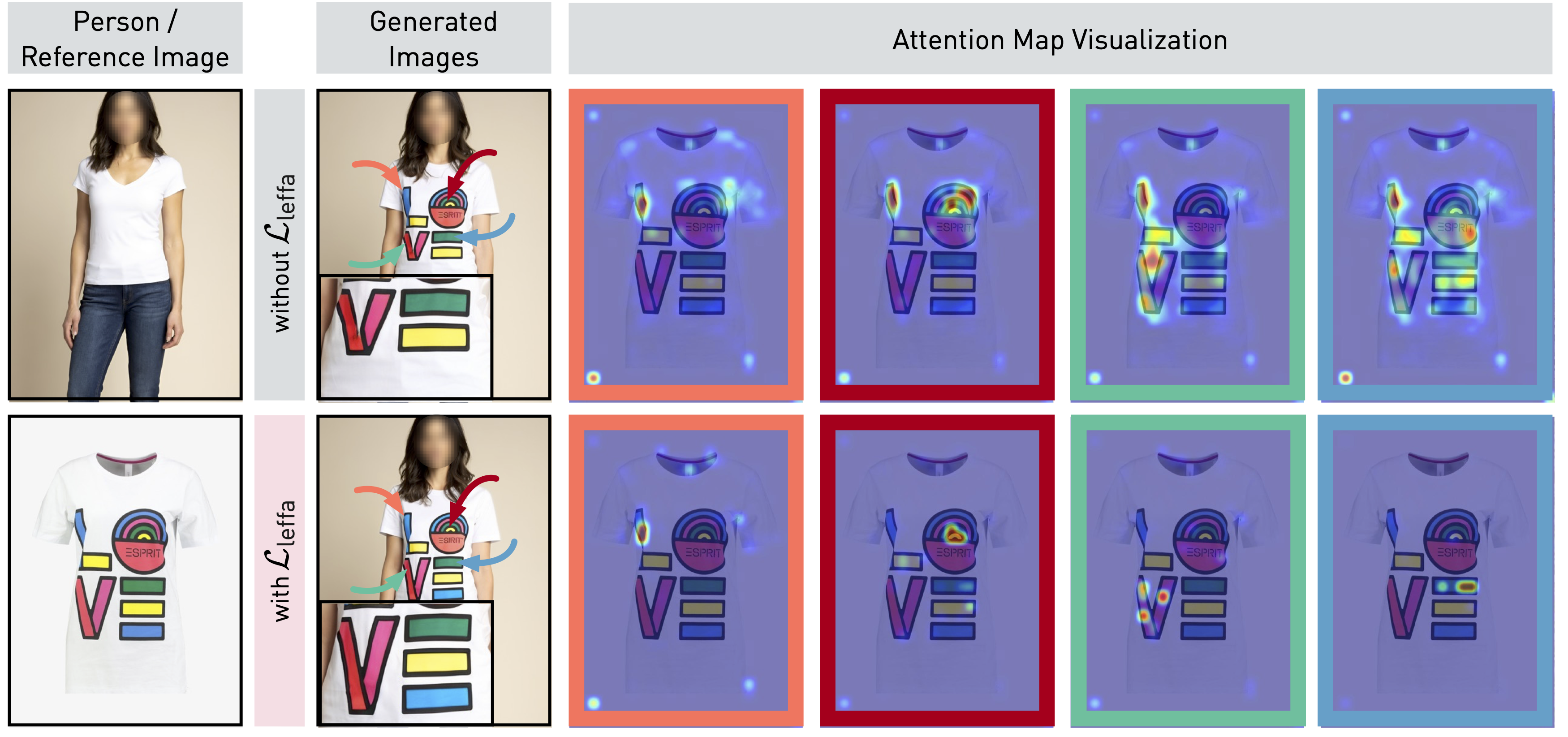}
    \vspace{-0.5cm}
    \caption{
    Visualization of feature maps to assess the impact of our \ours{} loss $\mathcal{L}_\text{leffa}$.
    With our \ours{} loss added, our method not only maintains overall generation quality but also more accurately preserves fine-grained details.
    Additionally, attention map visualizations indicate that, with our loss, the target query focuses more precisely on the correct reference region.
    }
    \label{fig:vis_ablation}
\end{figure}

To further validate the effectiveness of our method, we conduct ablation experiments on the VITON-HD dataset.

\noindent \textbf{Effect of $\mathcal{L}_{\text{leffa}}$}.
We first ablate \ours{} loss to evaluate the benefits of learning flow fields in attention.
As shown in Tab.~\ref{tab:general}, all metrics degrade without $\mathcal{L}_{\text{leffa}}$.
Specifically, the FID worsens by +0.77/+0.86 in the paired/unpaired setting, clearly demonstrating the contribution of our \ours{} loss to improving overall generation quality.
We further present visualizations of attention maps using \ours{} in Fig.~\ref{fig:vis_ablation}, which clearly show that the attended regions (highlighted in red) accurately correspond to the desired reference regions.

We additionally conduct a list of ablations to search the optimal training hyperparameters for our \ours{} loss.

\noindent \textbf{Effect of Different $\lambda_{\text{leffa}}$.} 
In Fig.~\ref{fig:ablation}, we find that by increasing $\lambda_{\text{leffa}}$ from $0.0$ to $10^{-3}$ improves performance, but further increasing to a degradation in performance.
This suggests that while the \ours{} loss effectively guides attention, overly strong supervision can hinder generalization, ultimately degrading generation quality.


\noindent \textbf{Effect of Resolution Threshold $\theta_{\text{resolution}}$.}
In Fig.~\ref{fig:ablation}, we observe that the optimal performance appears when $\theta_{\text{resolution}}$ is within $1/32$ and $1/16$.
Otherwise, attention maps are either too large (resulting in too few attention layers) or too small to focus on the correct regions.


\noindent \textbf{Effect of Timestep threshold $\theta_{\text{timestep}}$.}
When the sampled timestep $t$ is large, the target query would include too much noise, making it harder for the model to recognize the correct reference key regions.
In Fig.~\ref{fig:ablation}, we observe that the optimal performance is achieved at $\theta_{\text{resolution}} = 500$.
Increasing it to $750$ or $1000$ results in degraded performance even compared to one without using $\mathcal{L}_{\text{leffa}}$.


\noindent \textbf{Effect of Temperature $\tau$.}
Adjusting the temperature $\tau$ controls the range of focus that the target query has on the reference key.
As shown in Fig.~\ref{fig:ablation}, we test temperatures from $0.1$ to $5.0$ and find optimal performance at $\tau = 2.0$.
At $\tau = 0.1$, the attention map becomes overly sparse, making optimization difficult; while at $\tau = 5.0$, the attention map is overly dense, making it difficult to focus on the correct regions and eventually resulting in a significant performance drop.


\begin{figure}[t]
    \centering
    \vspace{-2mm}
    \includegraphics[width=0.75\linewidth]{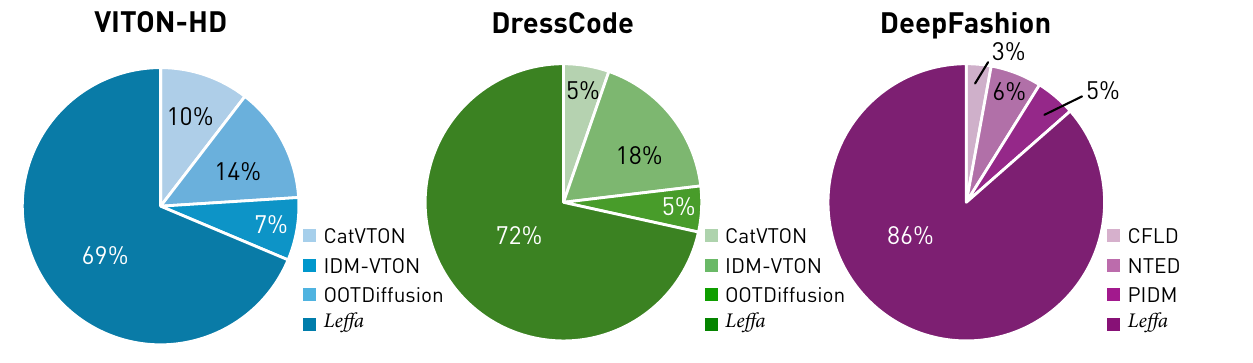}
    \vspace{-2mm}
    \caption{Preference percentage via human studies.
    Our \ours{} consistently outperforms previous methods across datasets.
    }
    \label{fig:human_study}
    \vspace{-4mm}
\end{figure}

\subsection{Human Study}
We have observed that some evaluation metrics, such as FID, are highly sensitive to implementation details~\citep{heusel2017gans, kynkaanniemi2022role, peebles2023scalable} and may not accurately reflect image generation quality.
To provide a more comprehensive evaluation, we conduct a human study comparing our \ours{} with other methods.

Specifically, VITON-HD~\citep{choi2021viton} and DressCode~\citep{morelli2022dress} are selected for virtual try-on , where we compare against OOTDiffusion~\citep{xu2024ootdiffusion}, IDM-VTON~\citep{choi2024improving}, and CatVTON~\citep{chong2024catvton}; while DeepFashion~\citep{liu2016deepfashion} is selected for pose transfer, comparing against PIDM~\citep{bhunia2023person}, NTED~\citep{ren2022neural}, and CFLD~\citep{lu2024coarse}.
We invite 50 participants to evaluate 90 cases (30 randomly selected cases from each dataset), with each participant selecting the image with the best generation quality from each test set.
As shown in Fig.~\ref{fig:human_study}, our method receives the highest number of preferred selections, significantly outperforming previous methods on both virtual try-on and pose transfer tasks.

\section{Related Work}
\label{sec:related_work}

\subsection{Controllable Person Image Generation}

\noindent \textbf{Appearance Control}.
Virtual try-on seeks to automate outfit changes in person images without distorting garment details, which is a longstanding challenge~\citep{yang2020towards, choi2021viton, ge2021disentangled, ge2021parser, he2022style}.
Recently, diffusion-based methods have emerged as the de facto solutions~\citep{zhu2023tryondiffusion, kim2024stableviton, xu2024ootdiffusion, wan2024improving}.
Some pioneering works~\citep{morelli2023ladi, gou2023taming, zeng2024cat}
enhance fine-grained detail preservation by incorporating additional modules, such as a textual inversion module with CLIP~\citep{radford2021learning} for enhanced conditions, a warping model to merge garments with garment-agnostic person images, and ControlNet~\citep{zhang2023adding} with DINOv2~\citep{oquab2023dinov2} to extract garment details.
Further improvements~\citep{kim2024stableviton, choi2024improving, xu2024ootdiffusion, wan2024improving} leverage complex model structure designs, external models and textual information for enhanced performance.
Recently, \citet{chong2024catvton} achieve detail preservation without extra models but requires complex training strategies (\eg, DREAM~\citep{zhou2024dream}) to reduce the training instability.

\noindent \textbf{Pose Control}.
Similarly, pose transfer presents a challenge in capturing the intricate structure of the pose transformation while preserving the fine-grained details of the textures~\citep{ma2017pose, siarohin2018deformable, li2019dense, zhu2019progressive, ren2020deep, men2020controllable, zhang2021pise, albahar2021pose, zhou2021cocosnet, sanyal2021learning, zhang2022exploring, zhou2022cross, ren2022neural}. 
\citet{bhunia2023person} first leverage diffusion models with a texture diffusion module for quality enhancement.
\citet{han2023controllable} further propose a pose-constrained attention module and replace skeletons with dense pose~\citep{guler2018densepose} for more accurate pose conditioning.
\citet{lu2024coarse} introduce a perception-refined decoder to progressively generate person images from coarse to fine while employing hybrid-granularity attention to enhance detail.
Recently, \citet{pham2024cross} replace UNet~\citep{ronneberger2015u} with the masked diffusion transformer~\citep{peebles2023scalable, gao2023masked}, achieving efficient and high-quality generation.

Although the existing methods have achieved significant progress on both virtual try-on and pose transfer tasks, most of them rely on complex module designs and extra models for fine-grained detail preservation.
In contrast, we introduce a method that ensures stable training without complex designs, reducing detail distortion without adding any inference cost or extra parameters.

\subsection{Attention in Controllable Generation}

The attention mechanism, a crucial module in generative models, has been extensively studied and shown to significantly enhance model generation quality when effectively guided~\citep{liu2024faster, xie2023boxdiff, phung2024grounded, guo2024focus, chefer2023attend, cao2023masactrl, xiao2024fastcomposer, hertzprompt}.
How to improve attention mechanism has also been studied in controllable person image generation.
Some methods~\citep{ren2020deep, zhou2021cocosnet, ren2022neural, tang2020xinggan} propose customized attention mechanisms.
For instance, \citet{ren2020deep} develop a differentiable global-local attention to reassemble inputs at the feature level, while \citet{zhou2021cocosnet} leverage iterative patch matching to enhance attention quality.
\citet{ren2022neural} apply neural texture extraction and distribution, using semantic filters to improve attention maps and image quality.
On the other hand, several methods directly optimize attention maps via supervision.
\citet{han2023controllable} propose a pose-constrained attention loss to model appearance-pose interactions.
\citet{kim2024stableviton} introduce an attention total variation loss to enhance focus on garment regions, though it lacks explicit guidance to attend to correct regions.
In contrast, we propose a regularization loss that guides attention maps to attend to correct regions without additional annotations, enhancing attention quality and reducing fine-grained detail distortion.

\section{Conclusion}
\label{sec:conclusion}
This paper introduces a regularization loss, learning flow fields in attention (\ours{}), to enhance controllable person image generation.
Our approach not only preserves high overall image quality but also mitigates fine-grained detail distortion.
We validate the effectiveness and generalization ability of \ours{} by integrating it with different diffusion-based methods, achieving significant qualitative and quantitative improvements in both virtual try-on and pose transfer tasks.
Future work will focus on developing a unified model that can simultaneously control appearance and pose.

\section{Limitation}
While \ours{} significantly improves controllable person image generation in appearance and pose control, it has several limitations. 
First, it requires multi-stage training with the \ours{} loss applied only in the final stage.
In future work, we aim to design a single-stage model to simplify the training process.
Second, appearance control relies on garment segmentation, which impacts performance when segmentation is inaccurate.
We plan to develop a mask-free approach to ensure high quality generation and preserve fine-grained details without distortion.
Third, our method struggles to preserve extremely fine-grained details, such as small text, due to the $8\times$ resolution compression brought by the latent encoder in the latent diffusion model.
It is worth noting that the issues mentioned above are not unique to our method but are also present in other methods.

\section{Statement}
All experiments, data collection, and processing activities were conducted at King's College London. Meta was involved solely in an advisory role and NO experiments, data collection or processing activities were conducted on Meta infrastructure.

\bibliographystyle{assets/plainnat}
\bibliography{main}

\clearpage
\newpage
\beginappendix

\appendix

In the supplementary material, we provide additional experimental details along with qualitative and quantitative results.
Additionally, we discuss our diffusion-based baseline and \ours{} loss, along with their limitations.

\section{Experimental Details}

\subsection{Datasets}
In this section, we provide a detailed introduction to the three datasets used in our study.

\noindent \textbf{VITON-HD}~\citep{choi2021viton} dataset is the most commonly used dataset for virtual try-on task.
The training set contains 11,647 person and garment image pairs, and the test set contains 2,032 pairs.
All images are front-view, upper-body garments with a resolution of $1024 \times 768$.

\noindent \textbf{DressCode}~\citep{morelli2022dress} dataset is composed of various types of garments, comprising 48,392 person and garment image pairs in the training set.
This includes 13,563 upper-body, 7,151 lower-body, and 27,678 dress garment pairs.
The test set contains 5,400 pairs, evenly distributed across 1,800 pairs each for upper-body, lower-body, and dress garments.
All images have a resolution of $1024 \times 768$.

\noindent \textbf{DeepFashion}~\citep{liu2016deepfashion} dataset includes high resolution 52,712 person images in the fashion domain.
Following ~\citet{zhu2019progressive}, we split the dataset into training and test subsets with 101,966 and 8,570 pairs, respectively.
Each pair includes the same person in the same garment but with different poses.

\section{More Results for DressCode Dataset}

To further validate our performance, we conduct separate evaluations on different garment categories for the DressCode dataset.
In Tab.~\ref{tab:dresscode_v2}, results show that our method significantly outperforms previous methods across all garment categories.
Specifically, for the upper body category, it achieves an FID reductions of -2.33 (paired) and -0.91 (unpaired); for the lower body category, -2.94 (paired) and -2.55 (unpaired); and for dresses, -1.25 (paired) and -1.71 (unpaired).

\begin{table}[!t]
\centering
\resizebox{0.75\linewidth}{!}{
\begin{tabular}{lcccccc}
\toprule[1.5pt]
\multirow{2}{*}{\textbf{Method}}        & \multicolumn{4}{c}{\textbf{paired}} & \multicolumn{2}{c}{\textbf{unpaired}} \\ \cmidrule{2-7} 
                                        & \textbf{FID} $\downarrow$ & \textbf{KID} $\downarrow$ & \textbf{SSIM} $\uparrow$ & \textbf{LPIPS} $\downarrow$ & \textbf{FID} $\downarrow$ & \textbf{KID} $\downarrow$ \\ \midrule
\rowcolor{tablegray} \multicolumn{7}{c}{upper body} \\
FS-VTON~\citep{he2022style}             & 11.29 & 3.65 & 0.941 & \underline{0.035} & 16.34 & 5.93 \\
HR-VITON~\citep{lee2022high}            & 15.36 & 5.27 & 0.916 & 0.071 & 16.82 & 5.70 \\
GP-VTON~\citep{xie2023gp}               & \underline{7.38}  & 0.74 & \underline{0.945} & 0.039 & 12.21 & 1.19 \\
LADI-VTON~\citep{morelli2023ladi}       & 9.53  & \underline{0.20} & 0.928 & 0.049 & 13.26 & 2.67 \\
DCI-VTON~\citep{gou2023taming}          & 7.47  & 1.07 & 0.942 & 0.041 & \underline{11.64} & \underline{0.86} \\
StableVITON~\citep{kim2024stableviton}  & 9.94  & 0.12 & 0.937 & 0.039 & -     & -    \\
OOTDiffusion~\citep{xu2024ootdiffusion} & 11.03 & 0.29 & -     & -     & -     & -    \\
\textbf{\ours{} (Ours)}                 & \textbf{5.05}  & \textbf{0.02} & \textbf{0.949} & \textbf{0.021} & \textbf{10.73} & \textbf{0.77} \\
\rowcolor{tablegray} \multicolumn{7}{c}{lower body} \\
FS-VTON~\citep{he2022style}             & 11.65 & 3.82 & 0.934 & 0.053 & 22.43 & 9.81 \\
HR-VITON~\citep{lee2022high}            & 11.41 & 3.20 & 0.937 & 0.045 & 16.39 & 4.31 \\
GP-VTON~\citep{xie2023gp}               & \underline{7.73}  & \underline{0.71} & 0.938 & \underline{0.042} & 16.70 & 2.89 \\
LADI-VTON~\citep{morelli2023ladi}       & 8.52  & 1.04 & 0.922 & 0.051 & \underline{14.80} & 3.13 \\
DCI-VTON~\citep{gou2023taming}          & 7.97  & 0.96 & \underline{0.939} & 0.045 & 15.45 & \underline{1.60} \\
\textbf{\ours{} (Ours)}                 & \textbf{4.79}  & \textbf{0.05} & \textbf{0.941} & \textbf{0.024} & \textbf{12.25} & \textbf{1.66} \\
\rowcolor{tablegray} \multicolumn{7}{c}{dresses} \\
FS-VTON~\citep{he2022style}             & 13.04 & 4.44 & \underline{0.888} & \underline{0.070} & 20.95 & 8.96 \\
HR-VITON~\citep{lee2022high}            & 16.82 & 4.89 & 0.865 & 0.113 & 18.81 & 5.41 \\
GP-VTON~\citep{xie2023gp}               & \underline{7.44}  & \textbf{0.32} & 0.881 & 0.073 & 12.64 & 1.83 \\
LADI-VTON~\citep{morelli2023ladi}       & 9.07  & 1.12 & 0.868 & 0.089 & 13.40 & 2.50 \\
DCI-VTON~\citep{gou2023taming}          & 8.48  & \underline{1.08} & 0.887 & \underline{0.070} & \underline{12.35} & \underline{1.36} \\
\textbf{\ours{} (Ours)}                 & \textbf{6.19}  & \textbf{0.32} & \textbf{0.891} & \textbf{0.044} & \textbf{10.64} & \textbf{0.59} \\
\bottomrule[1.5pt]
\end{tabular}
}
\vspace{-2mm}
\caption{Quantitative results comparison with other methods on the DressCode dataset for virtual try-on.
Our \ours{} achieves state-of-the-art results across all categories of garment.
}
\vspace{-2mm}
\label{tab:dresscode_v2}
\end{table}

\begin{table}[!t]
\setlength{\tabcolsep}{0.4cm}
\centering
\resizebox{0.75\linewidth}{!}{
\begin{tabular}{lccccccc}
\toprule[1.5pt]
\multirow{2}{*}{\textbf{Method}}  & \multirow{2}{*}{\textbf{$\mathcal{L}_{\text{leffa}}$}}        & \multicolumn{4}{c}{\textbf{paired}} & \multicolumn{2}{c}{\textbf{unpaired}} \\ \cmidrule{3-8} 
                                        & & \textbf{FID} $\downarrow$ & \textbf{KID} $\downarrow$ & \textbf{SSIM} $\uparrow$ & \textbf{LPIPS} $\downarrow$ & \textbf{FID} $\downarrow$ & \textbf{KID} $\downarrow$ \\ 
\midrule
\multirow{2}{*}{Ours}   & \xmark & 3.64  & 0.33 & 0.911 & 0.040 & 5.98  & 1.42 \\
                        & \cmark & \textbf{2.06}  & \textbf{0.07} & \textbf{0.924} & \textbf{0.031} & \textbf{4.48}  & \textbf{0.62} \\
\bottomrule[1.5pt]
\end{tabular}
}
\vspace{-2mm}
\caption{Ablation study on DressCode dataset for virtual try-on.
Our \ours{} loss significantly improves model performance.
}
\vspace{-2mm}
\label{tab:ablation_dress_code}
\end{table}

\begin{table}[!t]
\setlength{\tabcolsep}{0.7cm}
\centering
\resizebox{0.75\linewidth}{!}{
\begin{tabular}{lcccc}
\toprule[1.5pt]
\textbf{Method} & \textbf{$\mathcal{L}_\text{leffa}$} & \textbf{FID} $\downarrow$ & \textbf{SSIM} $\uparrow$ & \textbf{LPIPS} $\downarrow$ \\
\midrule
\multirow{2}{*}{Ours ($512 \times 352$)} & \xmark & 5.72 & 0.744 & 0.153 \\
                                    & \cmark & \textbf{4.23}  & \textbf{0.755} & \textbf{0.119} \\
\bottomrule[1.5pt]
\end{tabular}
}
\vspace{-2mm}
\caption{Ablation study on the DeepFashion dataset for pose transfer.
Our \ours{} loss significantly improves model performance.
}
\vspace{-2mm}
\label{tab:ablation_deep_fashion}
\end{table}

\section{More Ablation Studies}
Unless stated otherwise, the ablation studies are conducted on the VITON-HD dataset for the virtual try-on task, in alignment with the main paper.

\noindent \textbf{Effect of $\mathcal{L}_\text{leffa}$}.
To validate the effectiveness of \ours{} loss, we conduct ablation studies on the DressCode dataset for virtual try-on and the DeepFashion dataset for pose transfer.
i) As shown in Tab.~\ref{tab:ablation_dress_code}, our \ours{} reduces FID by -1.58 in the paired setting and -1.5 in the unpaired setting.
ii) Tab.~\ref{tab:ablation_deep_fashion} further shows an FID reduction of -1.49 on pose transfer task.
These results confirm that \ours{} loss significantly improves controllable person image generation, enhancing both appearance and pose control.

\begin{figure*}[t]
    \centering
    \includegraphics[width=0.95\linewidth]{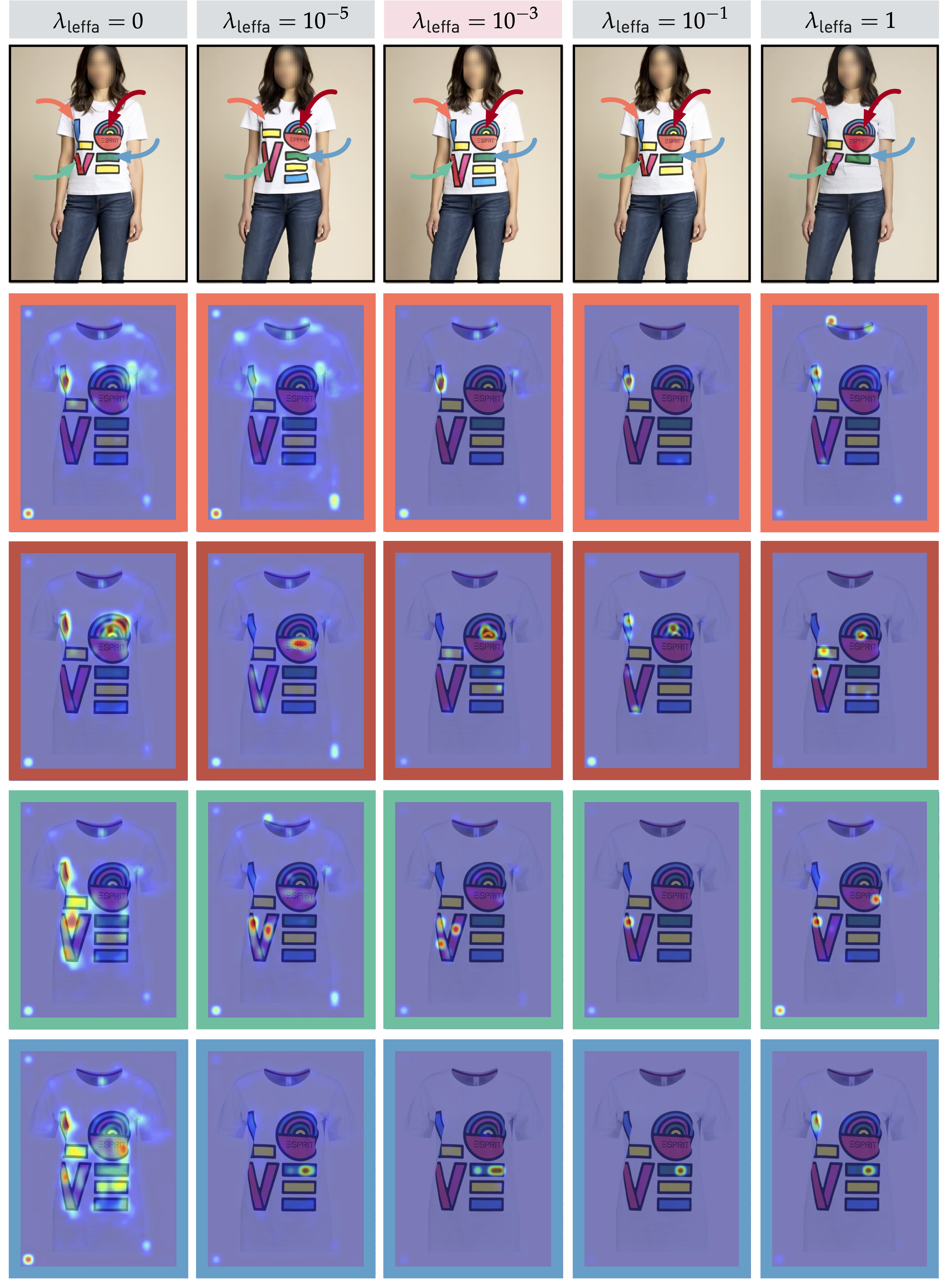}
    \vspace{-2mm}
    \caption{
    More visualizations of feature maps to assess the impact of \ours{} loss.
    The third column is the optimal setting used in our paper.
    }
    \label{fig:vis_ablation_v2}
\end{figure*}

\noindent \textbf{Qualitative impact of our \ours{} loss $\mathcal{L}_\text{leffa}$}.
In Fig.~\ref{fig:vis_ablation_v2}, we visualize attention maps with varying $\lambda_\text{leffa}$ to assess its impact on model training.
The first row shows generated results with different $\lambda_{\text{leffa}}$ values, with $\lambda_{\text{leffa}} = 0$ indicating no $\mathcal{L}_{\text{leffa}}$.
Rows 2 to 5 highlight reference key regions attended by the target query, marked by arrows of various colors.
Without $\mathcal{L}{\text{leffa}}$, attention is dispersed.
Increasing $\lambda_{\text{leffa}}$ improves focus, guiding the query to correct regions.
However, when $\lambda_{\text{leffa}} > 10^{-3}$, the attention becomes overly narrow and less accurate, hindering image generation.

\noindent \textbf{Effect of averaging attention map across multi-head}.
Inspired by~\cite{darcet2023vision}, we average the attention maps across all heads before computing the \ours{} loss.
As shown in Tab.~\ref{tab:more_ablation} (\ours{} w/o average $A$), computing the loss for each head individually degrades performance, emphasizing the role of redundancy in attention maps for better generalization.

\noindent \textbf{Effect of upsampling flow fields}.
To evaluate the impact of upsampling the flow fields to image resolution, we experiment with retaining their latent resolution for the \ours{}, requiring the ground truth image to be downsampled.
This resizing reduces detail, hindering accurate supervision.
As shown in Tab.~\ref{tab:more_ablation} (\ours{} w/o upsample $\mathcal{F}$), not upsampling the flow fields results in a performance drop, further supporting our viewpoint.

\begin{table}[!t]
\setlength{\tabcolsep}{0.4cm}
\centering
\resizebox{0.75\linewidth}{!}{
\begin{tabular}{lcccccc}
\toprule[1.5pt]
\multirow{2}{*}{\textbf{Method}}        & \multicolumn{4}{c}{\textbf{paired}} & \multicolumn{2}{c}{\textbf{unpaired}} \\ \cmidrule{2-7} 
                                        & \textbf{FID} $\downarrow$ & \textbf{KID} $\downarrow$ & \textbf{SSIM} $\uparrow$ & \textbf{LPIPS} $\downarrow$ & \textbf{FID} $\downarrow$ & \textbf{KID} $\downarrow$ \\ \midrule
\ours{} (Ours)                          & \textbf{4.54}  & \textbf{0.05} & \textbf{0.899} & \textbf{0.048} & \textbf{8.52}  & \textbf{0.32} \\
\ours{} w/o average $A$                 & 6.02 & 0.74 & 0.863 & 0.072 & 9.78 & 0.98 \\
\ours{} w/o upsample $\mathcal{F}$      & 4.94 & 0.32 & 0.888 & 0.064 & 9.33 & 0.78 \\
\bottomrule[1.5pt]
\end{tabular}
}
\vspace{-2mm}
\caption{Ablation study for the proposed \ours{} loss.}
\vspace{-2mm}
\label{tab:more_ablation}
\end{table}

\section{Discussion}
\label{sec:discussion}

\noindent \textbf{Why does our diffusion-based baseline perform comparably to state-of-the-art methods?}
Our baseline is similar to existing virtual try-on and pose transfer methods~\citep{choi2024improving, xu2024ootdiffusion, bhunia2023person, han2023controllable}, but we find that complex designs are unnecessary for strong performance.
The key lies in making both UNets in the dual architecture fully trainable, as freezing one significantly degrades results.
As shown in Tab.~\ref{tab:ablation_baseline}, freezing the reference UNet (as done in \citet{choi2024improving}) leads to a significant performance drop (FID reduction of -1.11/-1.25 for the paired/unpaired settings).
In contrast, adding CLIP visual and textual features~\citep{choi2024improving, xu2024ootdiffusion} results in only a slight performance decline.
This highlights that the key to improving the baseline lies in making both UNets trainable, while adding more complex designs (\eg, add CLIP, textual information) is unnecessary.

\begin{table}[!t]
\setlength{\tabcolsep}{0.4cm}
\centering
\resizebox{0.75\linewidth}{!}{
\begin{tabular}{lcccccc}
\toprule[1.5pt]
\multirow{2}{*}{\textbf{Method}}        & \multicolumn{4}{c}{\textbf{paired}} & \multicolumn{2}{c}{\textbf{unpaired}} \\ \cmidrule{2-7} 
                                        & \textbf{FID} $\downarrow$ & \textbf{KID} $\downarrow$ & \textbf{SSIM} $\uparrow$ & \textbf{LPIPS} $\downarrow$ & \textbf{FID} $\downarrow$ & \textbf{KID} $\downarrow$ \\ \midrule
Our baseline                & \textbf{5.31} & \textbf{0.30} & 0.885 & 0.058 & \textbf{9.38} & \textbf{0.91} \\
freeze Reference UNet       & 6.42 & 0.77 & 0.863 & 0.066 & 10.63 & 1.32 \\
+ CLIP visual feature       & 5.33 & 0.31 & \textbf{0.886} & \textbf{0.056} & 9.40 & 0.95 \\
+ CLIP textual feature      & 5.37 & 0.40 & 0.876 & 0.060 & 9.45 & 0.98 \\
\bottomrule[1.5pt]
\end{tabular}
}
\vspace{-2mm}
\caption{Ablation study for our diffusion-based baseline.
Making both the generative and reference UNets trainable is key to performance improvement for our diffusion-based baseline.
}
\vspace{-2mm}
\label{tab:ablation_baseline}
\end{table}

\noindent \textbf{Why not add $\mathcal{L}_{\text{leffa}}$ from the first training stage?}
At the start of training, the attention maps are not well learned, and applying our \ours{} loss too early forces the target query to prematurely attend to the reference key, hindering convergence rather than accelerating it.
Instead, introducing our \ours{} loss in a subsequent training stage significantly enhances performance, demonstrating its ability to correct inaccurate attention and guide the model toward more effective learning.

\end{document}